\documentclass[letterpaper, 10 pt, conference]{ieeeconf}

\IEEEoverridecommandlockouts                              
\usepackage{graphicx}
\usepackage{amsmath}
\usepackage{amssymb}
\usepackage{booktabs}
\usepackage{color, colortbl}
\usepackage{booktabs}
\usepackage[percent]{overpic}
\usepackage{xcolor}
\usepackage{verbatim}

\usepackage{multirow}

\usepackage{subcaption}
\usepackage{balance}

\usepackage[pagebackref,breaklinks,colorlinks]{hyperref}

\usepackage[capitalize]{cleveref}
\crefname{section}{Sec.}{Secs.}
\Crefname{section}{Section}{Sections}
\Crefname{table}{Table}{Tables}
\crefname{table}{Tab.}{Tabs.}

\newcolumntype{;}{!{\vrule width 2pt}}
\definecolor{LightYellow}{rgb}{1,1,0.7}

\definecolor{somegray}{rgb}{0.5, 0.5, 0.5}
\newcommand{\darkgrayed}[1]{\textcolor{somegray}{#1}}
\makeatletter
\newcommand*\titleheader[1]{\gdef\@titleheader{#1}}
\AtBeginDocument{%
  \let\st@red@title\@title
  \def\@title{%
    \vskip-3em
    \bgroup\normalfont\large\centering\@titleheader\par\egroup
    \vskip1.5em\st@red@title}
}
\makeatother

\titleheader{\darkgrayed{This paper has been accepted for publication at the \\
IEEE/RSJ International Conference on Intelligent Robots and Systems (IROS), Kyoto, 2022.
\copyright IEEE}}

\title{\LARGE \bf
Unsupervised confidence for LiDAR depth maps and applications
}

\author{Andrea Conti, Matteo Poggi, Filippo Aleotti and Stefano Mattoccia \\% 
University of Bologna%
}

\begin{document}

\twocolumn[{
\renewcommand\twocolumn[1][]{#1}
\maketitle
\begin{center}
    \begin{tabular}{c c c c}
        \begin{overpic}[height=0.12\textwidth]{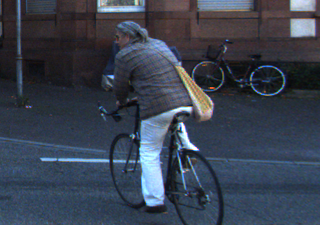}
            \put (4,6) {$\displaystyle\textcolor{white}{\textbf{(a)}}$}
        \end{overpic} &
        \begin{overpic}[height=0.12\textwidth]{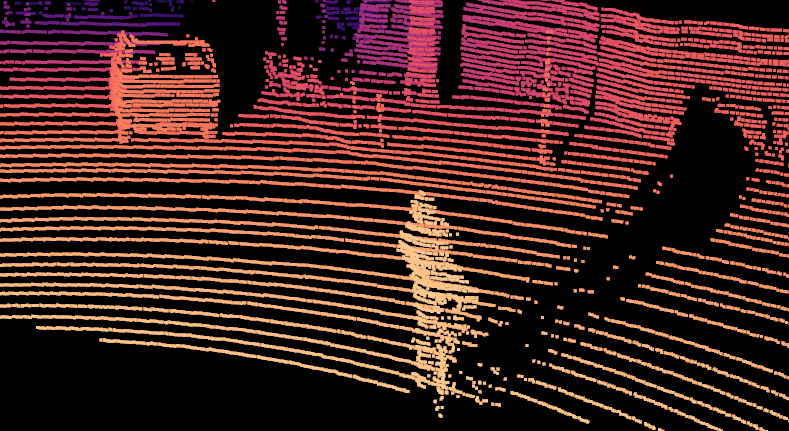}
            \put (4,6) {$\displaystyle\textcolor{white}{\textbf{(b)}}$}
        \end{overpic} &
        \begin{overpic}[height=0.12\textwidth]{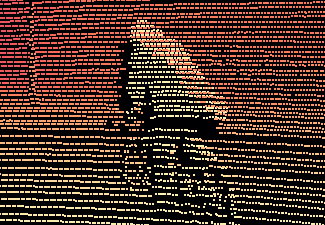}
            \put (4,6) {$\displaystyle\textcolor{white}{\textbf{(c)}}$}
        \end{overpic} &
        \begin{overpic}[height=0.12\textwidth]{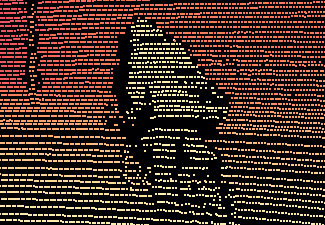}
            \put (4,6) {$\displaystyle\textcolor{white}{\textbf{(d)}}$}
        \end{overpic} \\
    \end{tabular}
    \label{fig:teaser}
\end{center}
\small \hypertarget{fig:teaser}{Figure 1.} \textbf{Outliers filtering in LiDAR depth maps}. Given an image (a) and a LiDAR pointcloud (b), the projection of the latter over the image plane does not properly handle occlusions between the two points of view, assigning wrong depth values to the foreground (c). Our method (d) learns to remove these outliers reliably and without supervision.
\vspace{0.2cm}
}]

\begin{abstract}

Depth perception is pivotal in many fields, such as robotics and autonomous driving, to name a few. Consequently, depth sensors such as LiDARs rapidly spread in many applications. The 3D point clouds generated by these sensors must often be coupled with an RGB camera to understand the framed scene semantically. Usually, the former is projected over the camera image plane, leading to a sparse depth map. Unfortunately, this process, coupled with the intrinsic issues affecting all the depth sensors, yields noise and gross outliers in the final output.
Purposely, in this paper, we propose an effective unsupervised framework aimed at explicitly addressing this issue by learning to estimate the confidence of the LiDAR sparse depth map and thus allowing for filtering out the outliers. Experimental results on the KITTI dataset highlight that our framework excels for this purpose. Moreover, we demonstrate how this achievement can improve a wide range of tasks.

\end{abstract}

%-------------------------------------------------------------------------
\section{Introduction}
\label{sec:introduction}

Depth perception plays a crucial role in computer vision, enabling it to tackle tasks such as autonomous driving, object manipulation, and more. The 3D structure of a sensed environment can be inferred through passive and active sensing technologies. The former has been deployed for decades using stereo \cite{scharstein2002taxonomy,poggi2021synergies}, structure-from-motion \cite{schonberger2016structure} or multi-view-stereo \cite{seitz2006comparison}. Each of these methods has its flaws and constraints. For instance, stereo depth perception requires two calibrated cameras and struggles where the scene lacks texture. 
On the other hand, active sensing relies on specialized sensors, and in the case of LiDARs (Light Detection And Ranging), flooding the scene with a laser beam and computing the distance of each point by measuring the traveling time of the ray. Despite being accurate, LiDARs struggle, for instance, when sensing not Lambertian surfaces due to multi-path interference or subsurface scattering. Moreover, the resulting point cloud is sparse and not coupled with any visual information. Thus, it is common to jointly use it with a standard camera and project the LiDAR point cloud over the camera image plane, resulting in a sparse depth map. However, this procedure raises a fundamental issue due to the different points of view of the two devices and the intrinsic sparsity of the LiDAR's output. Specifically, it leads to wrong depth values in the final RGB-D image, as shown in Figure \hyperref[fig:teaser]{1}. Therefore, \textit{LiDAR depth map filtering} is a serious problem to be tackled. 

To date, LiDAR sensors are massively used to source ground-truth data in primary scientific datasets, such as KITTI \cite{Geiger2013IJRR}, DrivingStereo \cite{yang2019drivingstereo}, and many others \cite{Argoverse, wang2019apolloscape, Gehrig21ral}, powering state-of-the-art deep learning techniques in computer vision. However, when projecting the depth map over the camera image plane, the issue mentioned above is usually tackled by enforcing consistency between the depth map and the values obtained through a stereo algorithm \cite{HirschmullerHeiko, uhrig2017sparsity} or deep stereo network \cite{yang2019drivingstereo}. Nonetheless, these approaches have some flaws as well: i) they require stereo cameras during acquisition with the LiDAR and ii) they do not filter out only the LiDAR errors, but also the stereo algorithm errors, thus affecting the cleaned data with the intrinsic limitations of the stereo setup (e.g., filtering out depth measures in textureless areas). Despite these limitations, these approaches are viable when pre-processing a dataset beforehand is feasible. However, they might not be applicable in real applications where a stereo setup is unavailable or when the depth labels for training are needed at runtime, for instance, to adapt stereo networks online \cite{Poggi2021continual}. 

To address all of these limitations, we propose a fast deep neural network framework, trained in an unsupervised manner, capable of predicting accurately the \textit{uncertainty} of the projected LiDAR sparse depth map using a simple RGB-D setup. Such uncertainty, or complementary \textit{confidence}, can be then deployed to filter out the errors, for instance, by enforcing a percentile to be removed or by using an absolute threshold. Therefore, the main novelties introduced by our work can be summarized as follows:

\begin{itemize}
    \item We propose the first deep learning framework designed to compute the confidence of LiDAR depth maps
    \item We provide a peculiar supervision scheme enabling an unsupervised training of the model, thus not requiring any expensive ground-truth depth annotation
\end{itemize}
Moreover, experimenting over two splits of KITTI \cite{Geiger2013IJRR} i) we uphold our claims assessing the effectiveness of our method versus existing alternatives \cite{ZhaoYiming_IEEE_2021,Eldesokey_2020_CVPR}, constantly outperforming even supervised techniques \cite{Eldesokey_2020_CVPR} and ii) we illustrate how filtering LiDAR depth maps with our approach yields consistent improvements in applications such as depth completion \cite{ma2018sparse,ma2019self}, guided stereo \cite{poggi2019guided} and sensor-guided optical flow \cite{Poggi_ICCV_2021} frameworks.
\section{Related work}
\label{sec:related work}

\textbf{LiDAR sensors in computer vision.}
The massive diffusion of LiDAR sensors in computer vision has begun with the release of the KITTI dataset \cite{Geiger2013IJRR}, an extensive collection of several thousand images and pointclouds acquired by a moving car equipped with a Velodyne LiDAR and stereo cameras. 
Eventually, more datasets followed this seminal work, such as DrivingStereo \cite{yang2019drivingstereo}, Argoverse \cite{Argoverse}, Apolloscape \cite{wang2019apolloscape} and DSEC \cite{Gehrig21ral}.
In between, over the KITTI dataset many computer vision tasks have been tackled, such as LiDAR SLAM \cite{wang2020intensity,pan2021mulls}, 3D object detection \cite{deng2020voxel,zheng2021se}, semantic segmentation \cite{behley2019iccv,cheng2021s3net}, object tracking \cite{wu2021,Kim21ICRA}, 3D scene flow \cite{rishav2020deeplidarflow,gojcic2021weakly3dsf}, fusion of LiDAR measurement with stereo \cite{poggi2019guided,Wang2019lidarstereo,cheng2019noise} or optical flow \cite{Poggi_ICCV_2021} and depth completion \cite{uhrig2017sparsity,ma2018sparse,ma2019self,Eldesokey_2020_CVPR,hu2020PENet}. The latter is one of the iconic problems in this field and processes LiDAR pointclouds projected into depth maps over the image plane as input. The standard benchmark for this task is hosted by KITTI \cite{uhrig2017sparsity} and has been obtained in a semi-automatic manner from Velodyne raw data by accumulating multiple point clouds and filtering outliers/moving objects employing consistency with a stereo algorithm \cite{HirschmullerHeiko}. As a result, ground-truth maps of the completion benchmark miss several labels where stereo algorithm struggle.

\textbf{Confidence estimation in computer vision.}
Estimating the confidence (or, complementary, the uncertainty) has been object of study in classical computer vision problems, such as optical flow \cite{sun2010secrets} or stereo \cite{scharstein2002taxonomy}. For the former task, we can distinguish between \textit{model-inherent} methods \cite{bruhn2006confidence,kybic2011bootstrap,wannenwetsch2017probflow}, that are part of the flow estimation model itself, and \textit{ad-hoc} approaches \cite{mac2012learning,kondermann2008statistical} that process already estimated optical flow maps. For the latter task, a similar distinction can be found in the literature, with methods processing the cost volume \cite{hu2012quantitative} or already estimated disparity maps \cite{poggi2017quantitative,poggi2021confidence}.
In deep learning, uncertainty estimation has been studied as well \cite{kendall2017uncertainties} and applied to specific tasks such as optical flow \cite{ilg2018uncertainty}, single image depth estimation \cite{Poggi_CVPR_2020}, visual odometry \cite{costante2020uncertainty}, semantic segmentation of LiDAR pointclouds \cite{cortinhal2020salsanext} and multi-task learning \cite{kendall2018multi} as well. 
However, prior works, in general, require some supervision from additional data, such as ground-truth labels for the specific task over which uncertainty is modeled \cite{ilg2018uncertainty} or additional images in case of self-supervision \cite{Poggi_CVPR_2020}. In contrast, in our case, we source supervision from the raw input data alone.

\begin{figure}[t]
    \centering
    \includegraphics[trim=0cm 14.25cm 15.25cm 0.10cm,clip,width=0.5\textwidth]{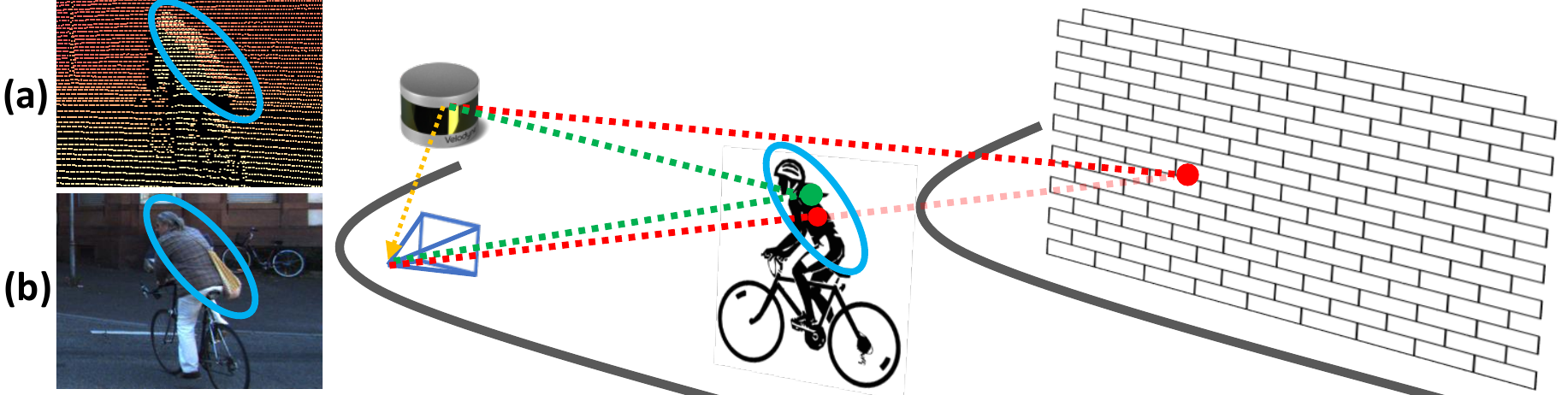}
    \caption{\textbf{Outliers formation process due to occlusions.}
    When a LiDAR and an RGB camera acquire from different viewpoints, projecting the point cloud into a depth map (a) on the image (b) introduces outliers (blue oval), e.g. points visible by the LiDAR occluded to the camera (red), yet projected near foreground points visible to both (green).
    }
    \label{fig:occlusions}
\end{figure}

\textbf{LiDAR confidence estimation.}
Few works allow computing the confidence on the LiDAR sparse depth map itself. Indeed, most attention to confidence estimation was given to estimating the uncertainty of the depth predicted by the overall framework using the LiDAR data, as in the case of depth completion \cite{eldesokey2019confidence, Eldesokey_2020_CVPR}. To the best of our knowledge, only the following works estimate the confidence on the input LiDAR sparse depth map. Eldesokey et al. \cite{Eldesokey_2020_CVPR} were the first explicitly modeling such confidence, but this happens as a side effect of making their depth completion framework more robust versus input noise. Thus it requires a massive amount of labeled data (i.e., the whole KITTI dataset with accumulated ground-truth depth maps) to train the model for its primary task: performing depth completion. Zhao et al. \cite{ZhaoYiming_IEEE_2021} propose instead a depth completion method that does not rely on learning, by using the local surface geometry of depth points, and enhance their system by employing a binary outliers detection algorithm.
Our solution differs from this latter since i) it is a learned method and ii) it generates confidence in place of a binary score, which allows for a finer outliers filtering mechanism, as we will see in our experiments.
Finally, concerning LiDAR filtering through a stereo setup, we mention the work by Cheng et al. \cite{cheng2019noise}, filtering the outliers in LiDAR depth maps while performing a noise-aware fusion with stereo data by checking consistency with the output of the fusion itself. 
\section{Proposed approach}
\label{sec:method}

At first, we introduce the reasons which give rise to outliers in LiDAR depth maps, and then we describe our framework specifically designed to filter them out. 

\subsection{Outliers in LiDAR depth maps}

There exist two leading causes of errors in LiDAR depth maps: i) erroneous measurements consequence of the LiDAR technology, for instance, originated by reflective or dark surfaces -- over which the behavior of the emitted beams become unpredictable -- or by other technological limitations (for instance, the mechanical rotation performed by the Velodyne HDL-64E used in KITTI \cite{geiger2012we}) and ii) incorrect projection of depth values near object boundaries, due to occlusions originated by the different viewpoints of the LiDAR sensor and the RGB camera.

Figure \ref{fig:occlusions} provides an intuitive overview about the second issue: concerning an urban scene acquired by a LiDAR and a camera, with a cyclist in the foreground and a wall far in the background (example available in the KITTI dataset). The different position of the two sensors causes some background regions to be visible to one of the two while occluded to the other. For instance, the red point on the wall is perceived by the LiDAR, but the cyclist occludes it in the image acquired by the camera. On the other hand, regions in the foreground are visible to both sensors, as the green point on the cyclist. When projecting LiDAR points into a depth map, specifically by mapping them over the camera image plane, depth values from occluded points in the background are projected into 2D pixel coordinates of foreground regions. The sparse nature of LiDAR points makes them visible in the resulting depth map shown in Figure \ref{fig:occlusions} (a), labeling the RGB image acquired by the camera (b) with wrong depth values in regions occluding the background points sensed by the LiDAR.

This bleeding effect, agnostic to the sensor accuracy, occurs in all LiDAR-Camera setups, including the depth maps made available by the KITTI completion dataset \cite{uhrig2017sparsity}, thus affecting the methods competing over the completion benchmark itself. Therefore, we introduce a carefully designed deep learning framework to deal with this issue, only using the RGB image coupled with the LiDAR depth.

\subsection{Architecture}

\begin{figure}[t]
    \centering
    \includegraphics[trim=1cm 14cm 13.5cm 0cm,clip,width=0.5\textwidth]{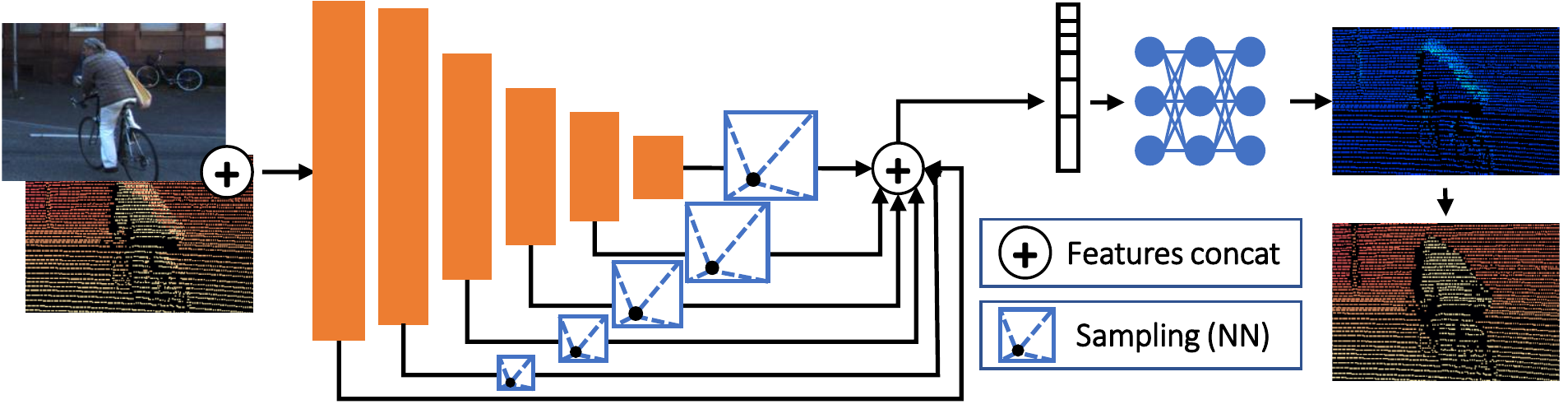}
    \vspace{-0.2cm}\caption{\textbf{Proposed architecture.} A convolutional encoder (orange) extracts features at different resolutions. We query features for each pixel with a valid LiDAR value and concatenate them (+) in a vector, fed to an MLP (blue) to estimate confidence.} 
    \label{fig:training_framework}
\end{figure}

Figure \ref{fig:training_framework} depicts the architecture designed to estimate confidence; it is composed of a multi-scale encoder and a prediction MLP (Multi Layer Perceptron) head. We feed the encoder with the concatenation of the RGB image and the sparse LiDAR depth map.

\textbf{Features extraction.}
The encoder of the network consists of three $3\times3$ conv2D layers with 32, 64 and 64 output channels, followed by five encoding blocks made by a $2\times2$ MaxPool operator with stride 2 and two $3\times3$ conv2D layers having the same number of output channels, respectively 128, 256, 512, 512 and 512 for the five blocks. The encoder extracts features at full resolution from the first three conv2D layers, and at five more resolutions from the five aforementioned blocks, respectively at $\frac{1}{2}$, $\frac{1}{4}$, $\frac{1}{8}$, $\frac{1}{16}$ and $\frac{1}{32}$. Extracting features at multiple scales allows increasing the receptive field and considering complex features from large areas in the image.  This strategy, for instance, allows evaluating the shape of a large complex object such as a car to tell which depth measurements are outliers. Furthermore, the multi-scale extraction plays a crucial role since we apply an MLP head that does not consider the local information around each feature vector due to its inherent nature. Leaky ReLUs follow each convolutional/fully connected layer.

\textbf{Confidence estimation.}
Once multi-scale features have been extracted, a sampling process occurs at each scale (using nearest-neighbor interpolation to sample at smaller scales) to compose a feature vector of size $64+128+256+512+512+512$ for each depth measurement contained in the LiDAR sparse depth map. The MLP head infers confidence by processing the high-level information regarding the image context and the original sparse depth distribution. Such an estimation comes in the form of variance, similarly to predictive uncertainty strategies \cite{kendall2017uncertainties} (the lower, the more confident). It is worth noting that a plain convolutional decoder could be used in place of an MLP. However, the specific task we are tackling does not require generating a dense output. Thus a simple MLP can estimate the confidence only for the meaningful pixels in the input depth map. We will show in Sec. \ref{sec:experimental results} how this approach steadily improves accuracy. 

\subsection{Unsupervised learning procedure}\label{subsec:unsupervised learning procedure}

The following section describes our peculiar unsupervised training procedure, not relying on any ground-truth data. 

\textbf{Proxy labels generation.}
To train our model, we argue that nearby pixels should share similar depth values \cite{poggi2016learning} except for points near discontinuities. Therefore, we take into account for each LiDAR depth point the other valid depth points inside a patch $P_N(x)$ of size $N \times N$ and compute a \textit{proxy label} representing a plausibly correct depth for each original LiDAR depth value available:

\begin{equation}
    d^*_x = f(\{d : d \in P_N(x), \ d > 0\})
\end{equation}

To speed up the training procedure and obtain a faster convergence, we use a fixed $f$ function. Precisely, we extract the minimum depth among the valid depths contained in the patch. Another approach might be to use the average of the valid depths in the patch. However, in the presence of occlusions, this strategy would cause both background and foreground depths to be detected as outliers since both are far from the average depth occurring between the two. In contrast, using the minimum depth value correctly selects the foreground points as reliable in the presence of occlusions. 
As a drawback, it may lead to indiscriminately detecting as outliers most of the pixels in the background, even if not occluded. 
However, in practice, we will show that the network learns to ameliorate this issue and that the patch size affects the performance to a lesser extent.
To support the effectiveness of our proposal, in Sec.  \ref{sec:experimental results} we compare the behaviour of the network using the minimum, the average, and the KITTI ground-truth depth itself as proxy labels.

\textbf{Loss function.}
We model the confidence of LiDAR depth $d$ assuming a Gaussian distribution centred in the proxy label $d^*$ with variance $\sigma^2$, the latter encoding the depth uncertainty. Thus, during training, the network learns to regress $\sigma$ by minimizing the negative log-likelihood of the distribution.

\begin{equation}\label{eq:nll_loss}
    \mathcal{L}_{G} = - \ln\left( \frac{1}{\sigma\sqrt{2\pi}} e^{-\frac{(d - d^*)^2}{2\sigma^2}}\right)
\end{equation}
We can rewrite (\ref{eq:nll_loss}) as follows:

\begin{align}
    \mathcal{L}_{G} & \approx \ln(\sigma) + \frac{(d - d^*)^2}{2\sigma^2} \label{eq:loss}
\end{align}

According to our experiments, (\ref{eq:loss}) becomes unstable when $\sigma \ll 1$ since it leads to enormous loss values hampering the learning procedure. Therefore, we constrain the network output to be $\sigma \ge 1$, obtaining the following additional advantages. The regularization term $\ln(\sigma)$ is 0 when $\sigma$ reaches its minimum value ($\sigma = 1$). Besides, small $\sigma$ values no longer magnify $(d - d^*)^2$, an unwelcome event since the network aims to minimise this term as much as possible. 

Nonetheless, by taking into account the derivative of (\ref{eq:loss}) 

\begin{equation}
    \frac{d}{d\sigma}\mathcal{L}_{G} = \frac{1}{\sigma} - \frac{(d - d^*)^2}{\sigma^3} \label{eq:loss_deriv}
\end{equation}
and solving for the minimum, we obtain $\sigma_{min} = |d - d^*|$. Hence, to constraint the minimum of the loss function in the chosen domain (i.e. $\sigma \ge 1$), we also need to enforce $(d - d^*)^2 \ge 1$ in (\ref{eq:loss}). Consequently, our final loss becomes: 

\begin{equation}
    \mathcal{L}_{G*} = \ln(\sigma) + \frac{(|d - d^*|+1)^2}{2\sigma^2},   \quad  \sigma \ge 1
    \label{eq:loss_final}
\end{equation}
In the next section we compare the performance of (\ref{eq:loss}), with $\sigma \ge 1$, and (\ref{eq:loss_final}) to measure the impact of this strategy.

Finally, it is worth observing that we might model uncertainty with other distributions such as the Laplacian, for which the previous observations still hold.  Additional details are available in the \textbf{supplementary material}. 
\section{Experimental results}
\label{sec:experimental results}

We now assess the effectiveness of our framework in comparison with state-of-the-art. 
Source code is available at \url{https://github.com/andreaconti/lidar-confidence}.

\begin{table}[t]
\centering
\scalebox{0.65}{
\begin{tabular}{cc}
\begin{tabular}{l ; c ; c ; c}
\\
Windows & {CV split} & {142 split} & {Average} \\
size & \cite{uhrig2017sparsity} & \cite{Menze2018JPRS} \\
\midrule
5$\times$5 & 0.1517 & 0.2291 & 0.1904\\
7$\times$7 & 0.1318 & \textbf{0.1975} & \underline{0.1647} \\
9$\times$9 & \textbf{0.1292} & \underline{0.1985} & \textbf{0.1639} \\
11$\times$11 & \underline{0.1316} & 0.1999 & 0.1658\\
13$\times$13 & 0.1372 & 0.2055 & 0.1714\\
\bottomrule
\end{tabular}
& \quad
\begin{tabular}{l ; c ; c }
Sampled & {CV split} & {142 split} \\
features & \cite{uhrig2017sparsity} & \cite{Menze2018JPRS} \\
\midrule
$\frac{1}{32}$ & 0.2436 & 0.3220 \\
$\frac{1}{32}$+$\frac{1}{16}$ & 0.1941 & 0.2597 \\
$\frac{1}{32}$+...+$\frac{1}{8}$ & 0.1680 & 0.2302 \\
$\frac{1}{32}$+...+$\frac{1}{4}$ & 0.1486 & 0.2109 \\
$\frac{1}{32}$+...+$\frac{1}{2}$ & \underline{0.1362} & \underline{0.2010} \\
All & \textbf{0.1292} & \textbf{0.1985} \\
\bottomrule
\end{tabular}\\
\textbf{(a)} & \textbf{(b)} \\
\end{tabular}}
\caption{\textbf{Experimental results -- ablation study.} We measure the impact of window size used to extract proxy labels $d^*$ and multi-resolution sampling. We report AUC values on the KITTI CV \cite{uhrig2017sparsity} and 142 \cite{Menze2018JPRS} splits. In each sub-table, best results are \textbf{bold} and second best \underline{underlined}.
}
\label{tab:ablation_study_windows_features}
\end{table}

\subsection{Evaluation dataset and training protocol}
\label{subsubsec:training and testing data}

We evaluate our framework on the KITTI dataset \cite{geiger2012we}, a standard benchmark in the field providing both images and raw LiDAR depth maps obtained from 151 video sequences, as well as accurate ground-truth labels. Such annotation, based on semi-automatic procedures, is highly time-consuming and requires stereo images.
For instance, the KITTI completion dataset \cite{uhrig2017sparsity} provides ground-truth maps obtained by accumulating 11 consecutive LiDAR pointclouds. Then, outliers (due to noise or moving objects) are removed by looking at inconsistency with respect to the output of the Semi-Global Matching (SGM) stereo algorithm \cite{HirschmullerHeiko}. This labeling strategy allows generating massive data (about 44.5K samples, 93K if considering stereo pairs) with the side-effect of losing several labels where LiDAR and SGM are not consistent. 
An even more accurate and laborious strategy consists of manually refining the labeling process, as done for the KITTI 2015 stereo dataset \cite{Menze2018JPRS}. In this case, 3D CAD models have been used to obtain an accurate annotation for cars at the cost of much more effort (indeed, only 200 annotated samples are available).

\begin{table}[t]
\centering
\scalebox{0.48}{
\begin{tabular}{cc}
\begin{tabular}{l l | l | l ; c ; c ; c}
\multicolumn{4}{c;}{} & {CV split} & {142 split} & {Average} \\
& Proxy & Head & Loss & {\cite{uhrig2017sparsity}} & {\cite{Menze2018JPRS}} \\
\midrule
& $d^*_{avg}$ & Decoder & $\mathcal{L}_{L}$ & 0.5286 & 0.8796 & 0.7041 \\
& $d^*_{avg}$ & Decoder & $\mathcal{L}_{L^*}$ & 0.2023 & 0.3730 & 0.2877\\
& $d^*_{avg}$ & Decoder & $\mathcal{L}_{G}$ & 0.3692 & 0.5064 & 0.4378\\
& $d^*_{avg}$ & Decoder & $\mathcal{L}_{G^*}$  & 0.1805 & 0.2403 & 0.2104\\
\midrule
$\ddag$ & $d^*_{avg}$ & MLP & $\mathcal{L}_{L}$ & 0.5890 & 0.9624 & 0.7757\\
        & $d^*_{avg}$ & MLP & $\mathcal{L}_{L^*}$ & 0.1565 & 0.2649 & 0.2107\\
$\ddag$ & $d^*_{avg}$ & MLP & $\mathcal{L}_{G}$ & 0.2430 & 0.2811 & 0.2621\\
        & $d^*_{avg}$ & MLP & $\mathcal{L}_{G^*}$ & 0.1546 & \underline{0.2197} & 0.1872 \\
\bottomrule
\end{tabular}
& \quad 
\begin{tabular}{ l l | l | l ; c ; c ; c }
\multicolumn{4}{c;}{} & {CV split} & {142 split} & {Average} \\
& Proxy & Head & Loss & {\cite{uhrig2017sparsity}} & {\cite{Menze2018JPRS}} \\
\midrule
$\ddag$ & $d^*_{min}$ & Decoder & $\mathcal{L}_{L}$ & 0.5569 & 0.7641 & 0.6605 \\
        & $d^*_{min}$ & Decoder & $\mathcal{L}_{L^*}$ & 0.1558 & 0.3693 & 0.2626\\
$\ddag$ & $d^*_{min}$ & Decoder & $\mathcal{L}_{G}$ & 0.8715 & 1.2760 & 1.0738\\
& $d^*_{min}$ & Decoder & $\mathcal{L}_{G^*}$ & 0.1382 & 0.2548 & 0.1965\\
\midrule
$\ddag$ & $d^*_{min}$ & MLP     & $\mathcal{L}_{L}$ & 0.6457 & 1.1370 & 0.8914\\
        & $d^*_{min}$ & MLP     & $\mathcal{L}_{L^*}$ & \textbf{0.1267} & 0.2446 & \underline{0.1857} \\
& $d^*_{min}$ & MLP     & $\mathcal{L}_{G}$ & 0.4801 & 0.6744 & 0.5773\\
& $d^*_{min}$ & MLP     & $\mathcal{L}_{G^*}$ & \underline{0.1292} & \textbf{0.1985} & \textbf{0.1639}\\
\bottomrule
\end{tabular}
\\
\end{tabular}}
\caption{\textbf{Experimental results -- ablation study.} We measure the impact of the three main design strategies in our unsupervised framework. We report AUC values on the KITTI CV \cite{uhrig2017sparsity} and 142 \cite{Menze2018JPRS} splits. In each sub-table, we report the best result in \textbf{bold} and the second-best \underline{underlined}. $\ddag$ means 10$^{-6}$ learning rate to avoid divergence.}
\label{tab:ablation_study}
\end{table}

In our experiments, we select two evaluation splits:
\begin{itemize}
 \setlength\itemsep{0.01em}
 \item \textbf{CV split}: composed of 1K images from the KITTI \underline{C}ompletion \underline{V}alidation set \cite{uhrig2017sparsity}
 \item \textbf{142 split}: a subset of 142 images from KITTI 2015 overlapping with KITTI completion, thus providing both raw LiDAR depth maps and manually annotated ground-truth
\end{itemize}
We train models using the 113 video sequences that do not overlap with any of the two splits. Moreover, since our framework quickly converges, we need just a few samples to achieve state-of-the-art results; thus, we use a subset of about 6K (one every five frames) samples. Nonetheless, for a fair comparison, we retrain our supervised competitor \cite{Eldesokey_2020_CVPR} over the whole available training set, yet avoiding overlapping with the 142 split (over which the weights released by the authors have been trained on).

Our framework is trained for 3 epochs only, using the ADAM optimizer with a learning rate of 10$^{-5}$,  with batches of 2 samples made of $320\times1216$ crops on a single NVIDIA RTX 3090. To train the models by Eldesokey et al. \cite{Eldesokey_2020_CVPR} we use the authors' code following the recommended settings.

\subsection{Outliers detection}

\textbf{Evaluation metrics.} We start by evaluating the performance of our method and existing approaches \cite{Eldesokey_2020_CVPR,ZhaoYiming_IEEE_2021} at detecting outliers in LiDAR depth maps. Purposely, we compute the Area Under the sparsification Curve (AUC), a standard metric for this task \cite{ilg2018uncertainty,poggi2021confidence,Poggi_CVPR_2020}. Namely, for each depth map in the dataset, pixels with both LiDAR and ground-truth depth available are sorted in increasing order of confidence score and gradually removed (2\% each time). The Root Mean Squared Error (RMSE) over remaining pixels is computed each time and a curve is drawn. The area under the curve quantitatively assesses the effectiveness at removing outliers (the lower, the better). Optimal AUC is obtained by removing pixels in decreasing order of depth error.

\begin{table}[t]
\centering
\scalebox{0.6}{
\begin{tabular}{cc}
\begin{tabular}{  l ; c ; c }
 & {CV split} & {142 split} \\
Method & {\cite{uhrig2017sparsity}} & {\cite{Menze2018JPRS}} \\
\midrule
Surface \cite{ZhaoYiming_IEEE_2021} & 0.7014 & 1.4446 \\
$|d-d^*_{avg}|$ & 0.2161 & \underline{0.2641} \\
$|d-d^*_{min}|$ & \underline{0.2053} &  0.2665 \\
\rowcolor{LightYellow}
Ours & \textbf{0.1292} & \textcolor{red}{\textbf{0.1985}} \\
\midrule
\rowcolor{white}
Optimal & 0.0271 & 0.0393 \\
\bottomrule
\multicolumn{3}{c}{\textbf{(a) unsupervised}} \\
\end{tabular}
& \quad\quad
\begin{tabular}{  l ; c ; c }
 & {CV split} & {142 split} \\
Method & {\cite{uhrig2017sparsity}} & {\cite{Menze2018JPRS}} \\
\midrule
NCNN-Conf-L1 \cite{Eldesokey_2020_CVPR} & \underline{0.1530} & \underline{0.3093} \\
NCNN-Conf-L2 \cite{Eldesokey_2020_CVPR} & 0.4624 & 0.8329 \\
pNCNN-Exp \cite{Eldesokey_2020_CVPR} & 0.8131 & 1.5430 \\
\rowcolor{LightYellow}
Ours \textdagger & \textcolor{red}{\textbf{0.1172}} & \textbf{0.2094} \\
\midrule
\rowcolor{white}
Optimal & 0.0271 & 0.0393 \\
\bottomrule
\multicolumn{3}{c}{\textbf{(b) supervised}} \\
\end{tabular}
\\
\end{tabular}}
\caption{\textbf{Experimental results -- outliers removal.} We report AUC values on the KITTI CV \cite{uhrig2017sparsity} and 142 \cite{Menze2018JPRS}, comparing with  unsupervised (a) and supervised (b) methods. Best method per category in \textbf{bold}, second best \underline{underlined}, absolute best in \textcolor{red}{\textbf{red}}. \textdagger{} means $d^*=$ ground-truth depth.}
\label{tab:auc}
\end{table}

\textbf{Ablation study.} We first assess the impact of several factors in our framework. In Table \ref{tab:ablation_study_windows_features}, we show the effect of the window size used to compute proxy labels $d^*$ and multi-resolution features sampling on our final model. For these experiments, we assume (\ref{eq:loss_final}) as loss function.
Table \ref{tab:ablation_study_windows_features} (a) shows that a 9$\times$9 patch allows us to train our framework at its best, while models trained on proxy labels computed on smaller or larger windows gradually achieve worse results. Intuitively, tiny windows lead the network toward over-fitting on high confidence values (i.e., more LiDAR values are likely to be close to $d^*$). While using larger windows leads to the opposite behavior (i.e., most LiDAR values will have a high difference compared to $d^*$ and drive, for instance, the network to predict low confidence in the presence of any depth discontinuity). Finally, Table \ref{tab:ablation_study_windows_features} (b) reports that, not surprisingly, the best results are obtained by sampling features from any resolutions, from full to $\frac{1}{32}$.

Then, we measure the impact of the different design choices used to implement our model. Specifically: i) different proxy label generation functions ($d^*_{min}$ and $d^*_{avg}$ for respectively the minimum and the average among the valid depths in the patch), ii) the prediction head (MLP or a five layers decoder with skip connections, where each layer has two $3\times3$ convolutional blocks followed by $2\times2$ nearest-neighbor upsampling) and iii) the distribution function underlying the loss term between Gaussian $\mathcal{L}_G$, Laplacian $\mathcal{L}_L$ \cite{kendall2017uncertainties} and the modified version of both $\mathcal{L}_{G^*}$ and $\mathcal{L}_{L^*}$ as described in Sec. \ref{subsec:unsupervised learning procedure}.
Table \ref{tab:ablation_study} collects results by several variants of our framework on both CV and 142 splits, using a 9$\times$9 window and sampling features at all resolutions following the outcomes from Table \ref{tab:ablation_study_windows_features}.
The scores are generally lower on the CV split because of the many missing labels from ground-truth maps obtained semi-automatically, resulting in several outliers being missing in the AUC evaluation. 
We can also notice that $\mathcal{L}_{L^*}$ and $\mathcal{L}_{G^*}$ always outperform their original counterparts, assessing the quality of our formulation. Moreover, the synergy between the minimum proxy label strategy and the MLP yields the best results overall. Finally, 
even if both $\mathcal{L}_{L^*}$ and $\mathcal{L}_{G^*}$ are competitive, we choose $\mathcal{L}_{G^*}$ since it leads to the best overall results. 

\begin{figure}
    \centering
    \renewcommand{\tabcolsep}{1pt}
    \begin{tabular}{ccccccc}
    \begin{overpic}[trim=17cm 0cm 16cm 3.5cm,clip,width=0.12\textwidth]{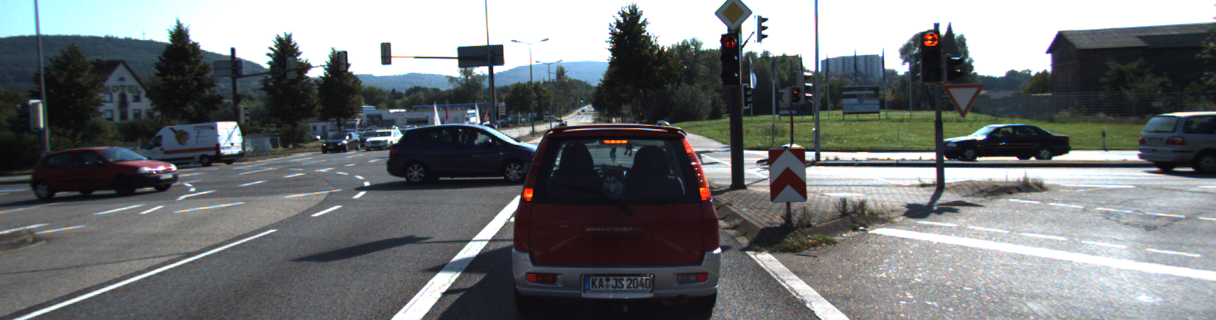} 
    \put (2,65) {$\displaystyle\textcolor{white}{\textbf{(a)}}$}
    \end{overpic} &
    \begin{overpic}[trim=17cm 0cm 16cm 3.5cm,clip,width=0.12\textwidth]{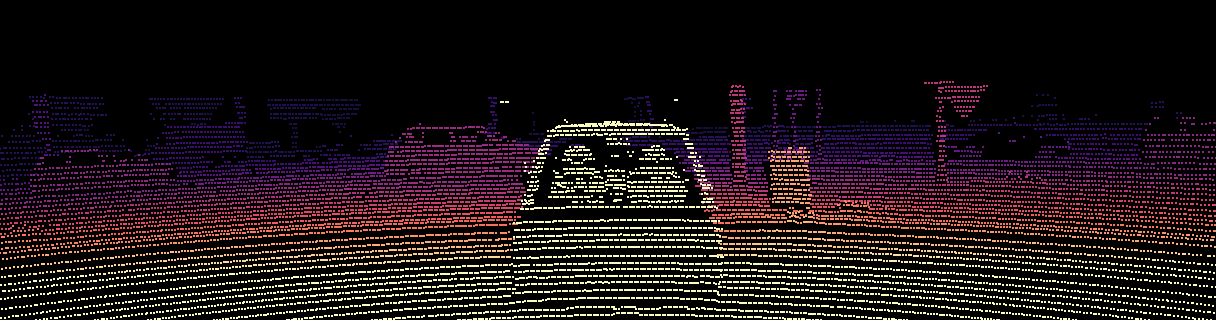} 
    \put (2,65) {$\displaystyle\textcolor{white}{\textbf{(b)}}$}
    \end{overpic} &
    \begin{overpic}[trim=17cm 0cm 16cm 3.5cm,clip,width=0.12\textwidth]{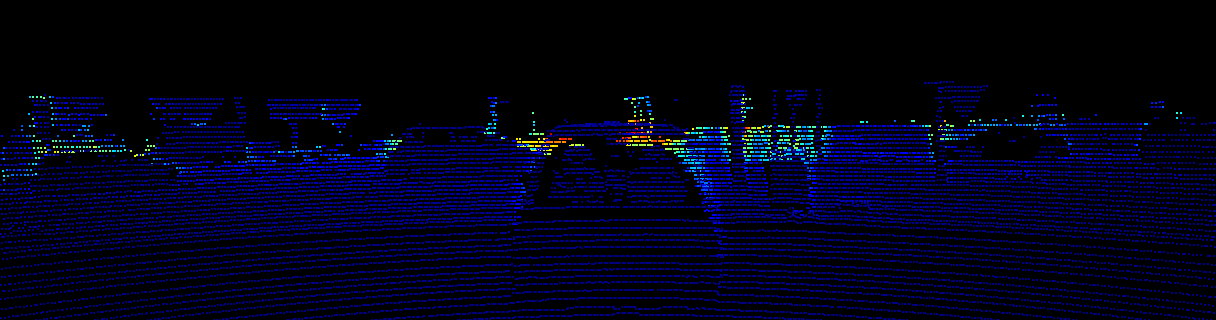}
    \put (2,65) {$\displaystyle\textcolor{white}{\textbf{(c)}}$}
    \end{overpic} 
    \\
    \end{tabular}
    \caption{\textbf{Qualitative results (142 split).} We show RGB images (a), raw LiDAR depth (b) and estimated confidence maps (c).}
    \label{fig:qualitatives}
\end{figure}

\begin{table}[t]
\centering
\scalebox{0.65}{
\begin{tabular}{cccc}
\begin{tabular}{  l ; r ; r }
 & {RMSE } & {\% filtered} \\
 & {(28.85\%} & {($\sim$0.68} \\
Filtering Method & filtering) & RMSE) \\
\midrule
SGM \cite{HirschmullerHeiko} & 0.6886 & 28.85 \\
\rowcolor{LightYellow}
Ours & \textbf{0.2085} & \textbf{3.90} \\
\midrule
\end{tabular}
&
\begin{tabular}{  l ; r ; r }
 & {RMSE } & {\% filtered} \\
 & {(20.08\%} & {($\sim$0.92} \\
Filtering Method & filtering) & RMSE) \\
\midrule
Reversing \cite{aleotti2020reversing} & 0.9171 & 20.08 \\
\rowcolor{LightYellow}
Ours & \textbf{0.2518} & \textbf{2.39} \\
\midrule
\end{tabular}
\\
\begin{tabular}{  l ; r ; r }
 & {RMSE } & {\% filtered} \\
 & {(1.47\%} & {($\sim$1.42} \\
Filtering Method & filtering) & RMSE) \\
\midrule
Surface \cite{ZhaoYiming_IEEE_2021} & 1.4189 & 1.47 \\
\rowcolor{LightYellow}
Ours & \textbf{1.1841} & \textbf{0.97} \\
\midrule
\end{tabular}
&
\begin{tabular}{  l ; r ; r }
 & {RMSE } & {\% filtered} \\
 & {(12.99\%} & {($\sim$0.72} \\
Filtering Method & filtering) & RMSE) \\
\midrule
LiDARStereoNet \cite{cheng2019noise} & 0.7176 & 12.99 \\
\rowcolor{LightYellow}
Ours & \textbf{0.3261} & \textbf{3.75} \\
\midrule
\end{tabular}
\\
\end{tabular}}
\caption{\textbf{Experimental results -- semi-automatic annotation.} We evaluate the annotation performance of our method on the 142 split, either by fixing the \% of filtered points or the final RMSE achieved by the competitors.} 
\label{tab:annotation}
\end{table}

\textbf{Comparison with state-of-the-art.} Table \ref{tab:auc} reports a comparison with existing approaches, namely Surface \cite{ZhaoYiming_IEEE_2021} and NCNN variants \cite{Eldesokey_2020_CVPR} on both splits, grouping unsupervised and non-learned methods on left (a) and supervised ones on right (b). 
Since our framework can be trained on ground-truth labels as well, we report this additional experiment, marked with $\dag$, to compare it with supervised methods directly. While this slightly improves the performance on the CV split, which is labeled with the same semi-automatic procedure of the training set, it leads to worse results on the accurate ground-truth maps of the 142 split. This outcome is not surprising since several outliers do not have a corresponding ground-truth value on KITTI CV and are never observed during supervised training over it. In contrast, as reported in the table, our unsupervised strategy is intrinsically unaffected by this bias.

Overall, our framework turns out the best approach, both when trained with and without ground-truth supervision. Indeed, even when trained in an unsupervised manner, it already outperforms supervised methods \cite{Eldesokey_2020_CVPR}. Moreover, we can notice that the absolute difference between LiDAR values and proxy labels is already a good cue to remove outliers, easily outperforming Surface \cite{ZhaoYiming_IEEE_2021} and often being better than supervised approaches \cite{Eldesokey_2020_CVPR}. Our framework learns to leverage such proxy labels and steadily exceeds their limitations, leading to even better results. Focusing on the former Table \ref{tab:auc} (a), we can notice how \cite{ZhaoYiming_IEEE_2021} performs poorly at sparsification. Indeed, Surface performs a binary classification of inliers and outliers, removing only a small set of pixels (respectively 1.60\% and 1.47\% of the pixels with available ground-truth on CV split and 142 split), yet leaving many outliers on stage. Nonetheless, since a binary method is penalized by AUC evaluation, we will provide more fair comparisons in Sec. \ref{sec:applications}.

Concerning supervised methods in Table \ref{tab:auc} (b), we interestingly notice that among NCNN variants, the one performing better at modeling uncertainty after completion according to \cite{Eldesokey_2020_CVPR}, i.e. pNCNN-Exp, is the worst at detecting outliers in the input. On the contrary, NCNN-Conf-L1 is the best variant on raw LiDAR -- although always outperformed by our approach, either supervised or unsupervised.

The superior accuracy achieved by our model comes at the cost of slightly higher complexity. Our network counts 16M weights versus the 300K of NCNN variants \cite{Eldesokey_2020_CVPR}, leading to higher runtime on both 3090 and Jetson TX2 GPUs -- respectively 0.02 and 1.02 seconds by our model versus 0.01 and 0.31 required by NCNN variants \cite{Eldesokey_2020_CVPR}. However, our model achieves better results and does not require any ground-truth depth label for training. For completeness, we also report the runtime required by Surface \cite{ZhaoYiming_IEEE_2021}. Although implemented on CPU, thus not directly comparable with our method and NCNN, \cite{ZhaoYiming_IEEE_2021} takes, respectively, 0.43 and 2.63 seconds on the same desktop PC equipped with a 3090 GPU -- and an i9-10900X -- and the Jetson TX2 CPU.

Figure \ref{fig:qualitatives} shows qualitative examples of confidence maps estimated by our unsupervised framework. More results are available in the \textbf{supplementary material}.

\subsection{Applications}\label{sec:applications}

Finally, we evaluate how our unsupervised model impacts some relevant applications making use of LiDAR data.

\textbf{Semi-automatic annotation.} The first direct application of our strategy consists of filtering LiDAR depth maps to obtain accurate, per-pixel depth annotations. Semi-automatic processes \cite{uhrig2017sparsity} usually rely on an external stereo setup and check for consistency between LiDAR values and disparity maps.
We compare our unsupervised model to this approach, either using a hand-crafted algorithm \cite{HirschmullerHeiko} or state-of-the-art self-supervised stereo networks \cite{aleotti2020reversing}, Surface \cite{ZhaoYiming_IEEE_2021} and a LiDAR-stereo fusion framework \cite{Wang2019lidarstereo}.
The comparison is performed on the 142 split since it provides manually annotated and refined ground-truth, in contrast to the CV split obtained semi-automatically. We limit to single depth map filtering and do not accumulate pointclouds over time to avoid issues with moving objects. When filtering using stereo methods, we convert LiDAR depth into disparity and filter pixels having a difference with the stereo disparity $>1$.

\begin{table}[t]
 \centering
 \scalebox{0.6}{ 
 \begin{tabular}{r ;lr;rrrr}
 \toprule
 & LiDAR & Removed & RMSE & MAE & iRMSE & iMAE \\
 Model & filtering & points (\%) & (mm) & (mm) & (1/km) & (1/km) \\
 \midrule
 Self-Sparse-to-Dense \cite{ma2019self} & None & None & 1102.062 & 303.007 & 4.316 & 1.670 \\
 Self-Sparse-to-Dense \cite{ma2019self} & Surface \cite{ZhaoYiming_IEEE_2021} & 3.74 & 974.443 & 295.587 & 4.313 & 1.665 \\
 \rowcolor{LightYellow}
 Self-Sparse-to-Dense \cite{ma2019self} & Ours & 1.20 & \textbf{959.100} & \textbf{288.457} & \textbf{4.187} & \textbf{1.640} \\
 \bottomrule
 \multicolumn{7}{c}{\textbf{(a)}} \\
 \toprule
 Sparse-to-Dense \cite{ma2018sparse} & None & None & 676.061 & 274.109 & 3.097 & 1.705 \\
 Sparse-to-Dense \cite{ma2018sparse} & Surface \cite{ZhaoYiming_IEEE_2021} & 3.74 & 703.597 & 276.701 & 3.087 & \textbf{1.689} \\
 \rowcolor{LightYellow}
 Sparse-to-Dense \cite{ma2018sparse} & Ours & 0.70 & \textbf{647.473} & \textbf{270.554} & \textbf{3.060} & 1.695 \\
 \bottomrule
 \multicolumn{7}{c}{\textbf{(b)}} \\
 \toprule
 PENet \cite{hu2020PENet} & None & None & 593.196 & 178.869 & 2.242 & 0.940 \\
 PENet \cite{hu2020PENet} & Surface \cite{ZhaoYiming_IEEE_2021} & 3.74 & 616.753 & 182.139 & 2.285 & 0.943 \\
 \rowcolor{LightYellow}
 PENet \cite{hu2020PENet} & Ours & 0.70 & \textbf{569.449} & \textbf{177.057} & \textbf{2.223} & \textbf{0.936} \\
 \bottomrule
 \multicolumn{7}{c}{\textbf{(c)}} \\
 \end{tabular}}
 \caption{\textbf{Experimental results -- Depth Completion.} Results on CV split by different completion models processing LiDAR filtered according to different strategies. Range: 50m.}
 \label{tab:completion50m}
\end{table}

In Table \ref{tab:annotation}, we report a sub-table for each competitor, measuring the percentage of pixels with both available LiDAR and ground-truth values that are discarded, as well as the filtered RMSE. The RMSE without filtering is 2.5698 meters. Since the four competitors rely on a binary criterion to remove outliers, we both commit to i) remove the same amount of pixels they do and prove that our framework better reduces the error, ii) reduce the RMSE to the same value as our competitors and prove that our framework can achieve such an error by removing fewer points.
Thus, in each comparison, we respectively i) remove the same percentage of the competitor and measure our final RMSE (first column), ii) filter pixels as long as we get the same RMSE of the competitor and measure the \% of pixels we removed to obtain it (second column).
Our method consistently achieves a much lower error when filtering the same percentage of pixels as the competitors. Moreover, it can reach the same final RMSE by removing a fraction of points, i.e. about 7-8 times less compared to stereo methods \cite{HirschmullerHeiko,aleotti2020reversing} yet not requiring two cameras as they do. Moreover, by committing to a single fixed threshold as one would do in a real application -- e.g., by constantly removing only 5\% pixels -- our model outperforms all the competitors, with 0.5850 RMSE. Finally, we can notice how leveraging stereo matching generally removes a high percentage of points (20-30\%) because of the several regions where stereo methods struggle, such as occlusions or untextured regions. In contrast, Surface \cite{ZhaoYiming_IEEE_2021} removes very few points but yields a significantly higher RMSE.

\textbf{Self-supervised/supervised depth completion.} We now show how filtering outliers improves the performance of networks for depth completion, the most iconic task performed starting from LiDAR depth maps, without specifically retraining neither our framework nor the depth completion network. Following \cite{ZhaoYiming_IEEE_2021}, Table \ref{tab:completion50m} shows results achieved by the Sparse-to-Dense framework -- using the weights released by the authors trained either without (a) \cite{ma2019self} or with (b) \cite{ma2018sparse} supervision -- when processing inputs that have been filtered through unsupervised techniques like ours and Surface \cite{ZhaoYiming_IEEE_2021}. We compute standard depth completion metrics, such as RMSE and Mean Absolute Error (MAE), inverse RMSE and inverse MAE on points up to 50m, to focus on the foreground objects (mostly affected by the outliers). Concerning the self-supervised variant (a), we can notice how filtering with Surface \cite{ZhaoYiming_IEEE_2021} improves all metrics by removing nearly 4\% of the total pixels with available LiDAR values. Concerning our method, we can achieve a larger improvement by limiting this percentage to 1.20\%, hinting that more precise filtering of the outliers, yet limited to fewer pixels, is more effective for the depth completion task. This is confirmed by experiments on the supervised variant (b): in this case, using Surface \cite{ZhaoYiming_IEEE_2021} only improves inverse metrics, while our method always improves all metrics by removing 0.70\% pixels only, resulting slightly worse only in iMAE compared to Surface.
{Finally, we report (c) experiments with PENet \cite{hu2020PENet}, a state-of-the-art framework for supervised completion, which confirm the previous findings.}

\begin{table}[t]
 \centering
 \scalebox{0.68}{
 \begin{tabular}{r ;lr;rrrr|r}
 \toprule
 & LiDAR & Removed &$>2$ & $>3$ & $>4$ & $>5$ & MAE \\
 Model & filtering & points (\%) & (\%) & (\%) & (\%) & (\%) & (px) \\
 \midrule
 PSMNet-ft-gd-tr & None & None & 5.14 & 3.39 & 2.69 & 2.29 & 1.08 \\
 \midrule 
 PSMNet-ft-gd-tr & Surface\cite{ZhaoYiming_IEEE_2021} & 4.37 & 4.75 & 3.01 & 2.34 & 1.95 & 1.01 \\
 \rowcolor{LightYellow}
 PSMNet-ft-gd-tr & Ours & 25.00 & \textbf{4.60} & \textbf{2.80} & \textbf{2.13} & \textbf{1.75} & \textbf{0.94} \\
 \bottomrule
 \end{tabular}}
 \caption{\textbf{Experimental results -- Guided Stereo.} Results on 142 split, with PSMNet weights provided by \cite{poggi2019guided} and guided with LiDAR filtered according to different strategies.}
 \label{tab:guided_stereo}
\end{table}

\textbf{Guided Stereo Matching.} We also measure the boost in performance of a sensor fusion framework combining passive stereo with LiDAR sensors filtering raw data with our method. Purposely, we choose the guided stereo framework \cite{poggi2019guided} (since it does not explicitly take into account noise, differently from \cite{cheng2019noise}) and collect in Table \ref{tab:guided_stereo} the accuracy yielded by PSMNet-ft-gd-tr -- the model provided by the authors -- on the 142 split and report the percentages of pixels with error larger than 2, 3, 4 and 5, together with MAE as in \cite{poggi2019guided}. While Surface \cite{ZhaoYiming_IEEE_2021} can slightly improve all metrics by removing less than 5\% of the total pixels with available LiDAR value, by filtering a more significant amount of pixels with our method, up to 25\%, we can further improve and achieve the best accuracy.
Interestingly, the guided stereo framework has an opposite behavior with respect to depth completion, as it benefits more from strict filtering.

\begin{table}[t]
\centering
\scalebox{0.65}
{
\begin{tabular}{c}
\begin{tabular}{l;lr;rr|rr}
\toprule
& LiDAR & Removed & EPE & Fl & \multicolumn{2}{r}{Density} \\
Guide Source & filtering & points (\%) & (px) & (\%) & \multicolumn{2}{r}{(\%)} \\
\midrule
Ego +RIC +MaskRCNN \cite{he2017maskrcnn} & None & None & 0.80 & 2.35 & \multicolumn{2}{r}{3.16} \\
Ego +RIC +MaskRCNN \cite{he2017maskrcnn} & Surface \cite{ZhaoYiming_IEEE_2021} & 4.37 &  0.74 & 2.35 & \multicolumn{2}{r}{3.06} \\
\rowcolor{LightYellow}
Ego +RIC +MaskRCNN \cite{he2017maskrcnn} & Ours & 7.00 &  \textbf{0.73} & \textbf{2.17} & \multicolumn{2}{r}{2.93} \\
\bottomrule
\end{tabular}
\\
\textbf{(a)}
\\
\begin{tabular}{l;lr;rr}
\toprule
& LiDAR & Removed & EPE & Fl \\
Model & filtering & points (\%) & (px) & (\%) \\
\midrule
guided-QRAFT \cite{Poggi_ICCV_2021} & None & None & 2.08 & 5.97 \\
\midrule
guided-QRAFT \cite{Poggi_ICCV_2021} & Surface \cite{ZhaoYiming_IEEE_2021} & 4.37 & 2.07 & 5.98 \\
\rowcolor{LightYellow}
guided-QRAFT \cite{Poggi_ICCV_2021} & Ours & 7.00 & \textbf{2.05} & \textbf{5.82} \\
\bottomrule
\end{tabular}
\\
\textbf{(b)}
\end{tabular}
}
\caption{\textbf{Experimental results -- Sensor-Guided Optical Flow.} Results on 142 split. (a) Accuracy of flow hints, obtained from LiDAR filtered according to different strategies, (b) accuracy of guided QRAFT \cite{Poggi_ICCV_2021} (CTK).}
\label{tab:sensorguidedflow}
\end{table}

\textbf{Sensor-Guided Optical Flow.} In the final evaluation, we leverage our framework to improve the performance of the Sensor-Guided Optical Flow pipeline \cite{Poggi_ICCV_2021}. It combines flow hints sourced using a LiDAR sensor with a deep optical flow network by filtering LiDAR points before hints computation. 
Table \ref{tab:sensorguidedflow} collects both the accuracy of flow hints (a) and the final results achieved by QRAFT weights trained on Chairs, Things and KITTI (CTK), as provided by the authors of \cite{Poggi_ICCV_2021}. On top, we can notice how Surface \cite{ZhaoYiming_IEEE_2021} can reduce the flow end-point error (EPE) of the computed hints, yet it cannot reduce the number of outliers with error larger than 3 pixels or 5\% (Fl). In contrast, by removing 7\% least confident pixels, our proposal can effectively improve all metrics.
At the bottom, we report the results achieved by guided QRAFT by using filtered hints. The impact of filtering is lower compared to other depth-related tasks. Indeed, in this task, LiDAR points are only one of several sources of errors, among camera pose estimation, flow estimation for dynamic objects and semantic segmentation. However, our method can consistently reduce both EPE and Fl metrics, whereas Surface cannot \cite{ZhaoYiming_IEEE_2021}.

\textbf{Current limitations.} 
Despite its effectiveness, we can see two main limitations in our framework. First, it adapts to the specific RGB + LiDAR setup used for training. Nonetheless, this appears to be a minor limitation since the training procedure is fast and completely unsupervised. The second concerns the need to choose a confidence threshold for outliers removal, which is however a concern shared with most approaches in this field \cite{ilg2018uncertainty,Postels_2019_ICCV,Poggi_CVPR_2020,poggi2021confidence,yang2020d3vo}.

\section{Conclusion}
\label{sec:conclusions}

In this paper, we tackle the nowadays common problem of detecting outliers in LiDAR depth maps obtained by projecting the pointclouds over the image plane of an RGB camera. To deal with it, we have proposed an unsupervised framework trained solely on an RGB image plus the raw LiDAR depth map to estimate the LiDAR sensor confidence. Compared to existing methodologies
\cite{ZhaoYiming_IEEE_2021,Eldesokey_2020_CVPR}, it yields state-of-the-art performance. Moreover, it constantly improves relevant tasks relying on LiDAR depth maps, such as semi-automatic annotation, depth completion \cite{ma2018sparse,ma2019self}, guided stereo \cite{poggi2019guided} and sensor-guided optical flow \cite{Poggi_ICCV_2021}.

\textbf{Acknowledgment.} This work was partially funded by University of Bologna and Ministero dello Sviluppo Economico (MISE) within the Proof of Concept 2020 program.

{\small
\bibliographystyle{ieee_fullname}
\bibliography{egbib}
}

\end{document}